\documentclass[10pt,twocolumn,letterpaper]{article}

\usepackage{iccv}
\usepackage{times}
\usepackage{epsfig}
\usepackage{graphicx}
\usepackage{amsmath}
\usepackage{amssymb}
\usepackage{algorithm2e,algorithmicx}
\usepackage{booktabs} % for much better looking tables
\usepackage{array} % for better arrays (eg matrices) in maths
\usepackage{paralist} % very flexible & customisable lists (eg. enumerate/itemize, etc.)
\usepackage{verbatim} % adds environment for commenting out blocks of text & for better verbatim
\usepackage{subfig} % make it possible to include more than one captioned figure/table in a single float
\usepackage[pagebackref=true,breaklinks=true,letterpaper=true,colorlinks,bookmarks=false]{hyperref}

\iccvfinalcopy % *** Uncomment this line for the final submission

\def\iccvPaperID{1702} % *** Enter the ICCV Paper ID here

\ificcvfinal\fi
\begin{document}

%%%%%%%%% TITLE
\title{Visual Word Selection without Re-Coding and Re-Pooling}

\author{Fatih Cakir ~ Stan Sclaroff\\
Department of Computer Science\\
Boston University\\
Boston, MA 02215 USA\\
{\tt\small \{fcakir,sclaroff@cs.bu.edu\}}
% For a paper whose authors are all at the same institution,
% omit the following lines up until the closing ``}''.
% Additional authors and addresses can be added with ``\and'',
% just like the second author.
% To save space, use either the email address or home page, not both
%\and
%Second Author\\
%Institution2\\
%First line of institution2 address\\
%{\small\url{http://www.author.org/~second}}
}

\maketitle
% \thispagestyle{empty}

%%%%%%%%% ABSTRACT
\begin{abstract}
The Bag-of-Words (BoW) representation is widely used in computer vision. The size of the codebook impacts the time and space complexity of the applications that use BoW. Thus, given a training set for a particular computer vision task, a key problem is pruning a large codebook to select only a subset of visual words. Evaluating possible selections of words to be included in the pruned codebook can be computationally prohibitive; in a brute-force scheme, evaluating each pruned codebook requires re-coding of all features extracted from training images to words in the candidate codebook and then re-pooling the words to obtain a representation of each image, e.g., histogram of visual word frequencies. In this paper, a method is proposed that selects and evaluates a subset of words from an initially large codebook, without the need for re-coding or re-pooling. Formulations are proposed for two commonly-used schemes: hard and soft (kernel) coding of visual words with average-pooling. The effectiveness of these formulations is evaluated on the 15 Scenes and Caltech 10 benchmarks.
\end{abstract}

\begin{figure}[t!]
  
  \centering
    \includegraphics[width=0.3\textwidth]{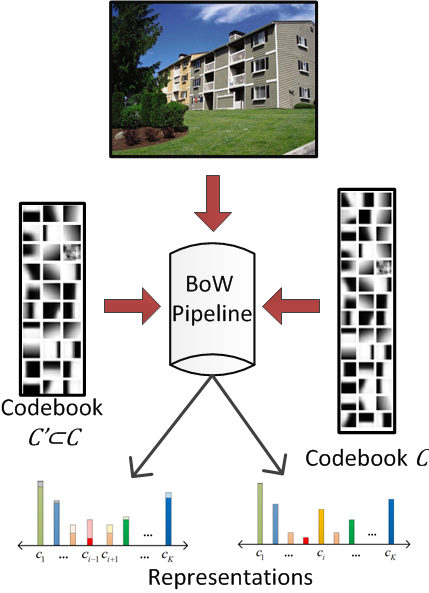}
    \caption{\label{fig:BoW_pipeline}To analyze the classification performance of possible subsets ${C' \subset C}$ of visual words requires instantiating the BoW pipeline to compute the new representation of an entire corpus with respect to ${C'}$. This is computationally very demanding, since it requires re-coding and re-pooling of all items in the corpus with respect to ${C'}$.  
%In this paper, we propose a codebook pruning method that avoids re-coding or re-pooling. 
}
\end{figure}

%%%%%%%%% BODY TEXT
\section{Introduction}

The Bag-of-Words approach (BoW) is now a standard image representation scheme employed in the computer vision community. A quick search on Google Scholar shows that approximately $20\%$ of the papers in the recent proceedings of the top three vision conferences contain the term \textit{``bag of words''}.\footnote{These conferences are ICCV 2011, ECCV 2012 and CVPR 2012.} The BoW pipeline (illustrated in Fig.~\ref{fig:BoW_pipeline}) comprises: extracting features, coding features with respect to a learned codebook, and pooling coded features to obtain the final representation of an image \cite{surka2004, lazebnik2006}.\footnote{In this paper, the following terms are used interchangeably:  codeword, visual word and basis; codebook and vocabulary; region and Voronoi cell.} 
The computational cost of the BoW pipeline  is usually dominated by the coding step, i.e., computing coding vectors corresponding to the extracted local features of an image. Coding is especially a bottleneck when the local features are sampled densely and the size of the codebook is kept large. 
%
%Another problem with large codebooks is related to storage. 
The dimensionality of the representation vector is usually a function of the cardinality of the codebook, and larger codebooks generally result in larger representation vectors and, thus, they require more storage.  Moreover, larger codebooks can in fact lead to degraded classification or retrieval accuracy, due to the curse of dimensionality \cite{stat2001}.  
% In fact, selecting a subset of codewords to include in the vocabularly can result in improved accuracy. %, as is confirmed in our experiments.

Codeword selection methods that find a subset of the vocabulary  that is most discriminative for a given task have been proposed to alleviate the abovementioned problems \cite{yang2007,triggs2005,winn2005,boostf, pair2010}. Many selection methods are adapted from the document retrieval domain, which uses criteria such as the term frequency, information gain, or ${\chi}^{2}$ measure to select terms to prune from the initial vocabulary with minimal sacrifice in retrieval/categorization accuracy \cite{Yang97acomparative}. 

However, there is a caveat: to the best of our knowledge all previous codeword selection schemes still require computing coding vectors with respect to the initial, larger vocabulary. Therefore, the prior work on this problem does not \textit{truly} reduce the size of the codebook in the sense that codewords deemed unworthy are not \textit{discarded}. Instead, previous methods generally use the full codebook to obtain an initial image representation, and then the reduced-dimensionality representation is computed from that. Consequently, if the initial vocabulary size is large then the computational cost of coding may still result in inefficiencies, especially on low performance platforms. Moreover this large vocabulary must still be retained in the system.
% After the selection process in previous approaches, computing coding vectors solely based on the selected subset of visual words cannot be considered since the BoW pipeline may then output totally different representation vectors nullifying the result of the initial codeword selection process. 

In order to truly reduce the size of the codebook, one must analyze the image representations computed under different subsets ${C'}$ of the visual words from the initial, larger codebook ${C}$ and select the best subset with respect to some given criteria. %This function can be similar to the criteria used in the aforementioned codebook reduction methods. 
One major drawback in this alternative scheme is the huge computational cost, since for each different subset of codewords ${C'}$ the BoW process must be instantiated in order to compute the new representations of an entire corpus.
\begin{figure}[t]
\begin{center}
 \includegraphics[width=0.95\linewidth]{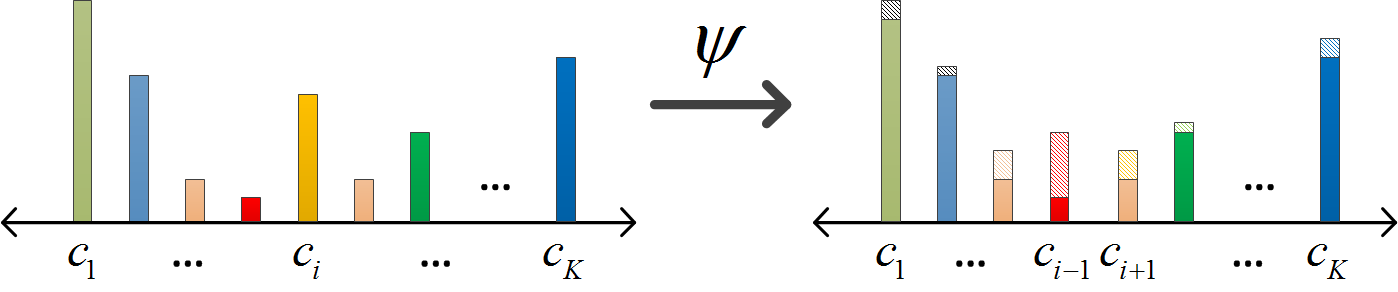}
\end{center}
\caption{When a visual word is omitted from the codebook, the representation vector (a histogram in this figure) is altered. 
%In past techniques, this requires computationally expensive re-coding and re-pooling. 
In the proposed formulation, when a visual word is omitted $\psi$ is used to approximate the feature vector, thereby avoiding re-coding and re-pooling.}
\label{fig:long}
\label{fig:onecol}
\end{figure}

In this paper, we provide a unique perspective on the BoW process that will allow us to compute the representations under subsets ${C'}$ very efficiently. Specifically, given an image representation vector computed with respect to a vocabulary ${C}$, we formulate a technique that can approximately infer the vector representation when a visual word is pruned from the vocabulary (illustrated in Fig.~\ref{fig:long}).  In this paper, we focus on assignment-based coding techniques, i.e., hard and soft (kernel) coding with average pooling, which together have wide adoption and are the basis for many other coding schemes in the literature \cite{sv2010,fish2010,vlad}.
%\footnote{In other words, under a certain case, we infer what would the representation vector be if rather than the full codebook, only a subset of it had been considered during the coding and %pooling steps.}, namely when the \textit{only knowns are this initial representation and the vocabulary itself.} 
Our observation is that, assuming the initial codebook construction step partitions the feature space allowing a generative model interpretation, one could use this structure to infer the alteration of a representation vector without the need for re-coding or re-pooling.  
 %The partitioning and generative model assumptions are not limited since the most popular schemes for building or describing codebooks like $K$-means and Gaussian Mixture Models (GMMs) allow us to make such assumptions. The alteration of the representation of an image when certain visual words are omitted also depends on the coding and pooling techniques being employed. Although there are numerous studies in this area, we analyze hard/soft coding in conjuction with average pooling since these are the most widely used schemes demonstrating state-of-the-art performances, in many vision applications. The overall analysis will then allow us to bypass the coding and pooling steps in computing the representation vector under different subsets of the visual words (Fig. 1). 

Based on our formulation, we demonstrate an efficient simulated annealing algorithm for decreasing the size of a codebook with respect to a classification task. We evaluate our algorithm on the 15-Scenes \cite{lazebnik2006} and Caltech-10 \cite{caltech101} benchmarks, and compare against two codeword selection solutions \cite{fulkerson,yang2007}. We demonstrate at least competitive classification performance at the gain of a decreased computational complexity in codebook pruning and decreased space complexity because we do not need to retain the initial codebook for use in coding new images. In summary, we make two primary contributions:
\begin{enumerate}
\item A method for inferring the representation vector for the hard and soft (kernel) coding methods, without doing coding or pooling in a BoW model when visual words are pruned from a vocabulary,

\item A codeword pruning scheme that eliminates the burden of considering the initial vocabulary in coding new images with respect to a reduced codebook. 
\end{enumerate}
The outline of the paper is as follows. In Section 2, we review related work. % regarding the Bag-of-Words model and codeword selection techniques. 
In Section 3, we describe our formulation. In Section 4, we present the experimental setup and discuss about the results. Finally, in Section 5 we provide concluding remarks and future work.

\section{Related Work }
Among many coding techniques in the literature we list only the most relevant and notable ones. These are methods like hard and soft (kernel) coding in which the features of an image are encoded by assigning them to the codebook entities \cite{sivic2003, soft2008}, methods that solve an optimization problem to determine the coding parameters \cite{sparse2009, llc2010}, and techniques that consider characterizing an image with the gradient information derived from a probability density function that models the generation process of the local features \cite{fish2010}. The computational cost is dominated by the coding step in these works, especially when the local features are sampled densely and the size of the codebook is  large.
%Although slight superiority of Fisher kernel based coding schemes has been demonstrated in the image classification domain \cite{devil2011}, conceptual easiness and near competitive performances of assignment based methods have enabled their wide adoption and also motivated research to improve their efficacy. 

%\subsection{Codeword Selection}
Borrowing ideas from the document retrieval domain \cite{Yang97acomparative}, traditional codeword selection methods use criteria such as the term frequency, 
${\chi}^{2}$ statistic, mutual information and learned SVM weights to select the most discriminative codewords \cite{yang2007, triggs2005}. Winn and Minka \cite{winn2005} propose to merge visual words/textons with respect to a probabilistic measure defined on the altered representations. Doing so they aim to find dimensions in the original representation to merge that presumably correspond to the same textures but are captured under different lighting or viewing angles. Similarly, Fulkerson et al. \cite{fulkerson} merges pairs of visual words based on a mutual information measure. Wang \cite{boostf} employs a boosting mechanism where each weak classifier is associated with a codeword and selection of weak classifiers in the procedure naturally results in the selection of the most discriminative codewords. Zhang, et al. \cite{zhang} considers an unsupervised scheme in which the visual words are selected by constructing a ridge regression formulation. 

Note that in all of these works, the requirement of computing the coding vectors with respect to the initial codebook still remains.

%------------------------------------------------------------------------

\begin{table}
\centering
\begin{tabular}{ c|p{7 cm}}
  \hline 
\hline                       
  $\mathbf{x} $ & local feature extracted from a local patch \\
  $\mathbf{V} $ & vocabulary (codebook) \\
  $K$ & cardinality of the vocabulary  \\
	$d$ & dimensionality of $\mathbf{x}$ \\
$\mathbf{c} $& codeword/visualword/basis of vocabulary \\
${\mathbf{h}}_{i}$ & coding vector of local feature ${\mathbf{x}}_{i}$\\
$\mathcal{F}$ & space of the coding vector \\ 
${\mathbf{f}}_{i}$ & representation vector after pooling coding vectors ${{\mathbf{h}}_{i}}$\\ 
$\underline{{\mathbf{h}}_{i}}$ & coding vector after pruning a basis from codebook\\
$\underline {\mathcal{F}}$ & dimensionality reduced coding vector space \\ 
$\underline{{\mathbf{f}}_{i}}$ & representation vector after pooling coding vectors $\underline{{\mathbf{h}}_{i}}$\\ 
$I$ & set of visual word indices in the codebook\\ 
$\mathcal{R}$ & a particular region or voronoi cell in the codebook space\\ 
$\underline{\mathcal{R}}$ & a particular region or voronoi cell in the codebook space when a basis is pruned\\ 
$y$ & a class/category\\ 
$N$ & Number of local features in an image\\ 
  \hline  
\hline
\end{tabular}
\caption{Notation.}
\end{table}

\section{Formulation}
\subsection{Mathematical Background}
The basic idea behind the Bag-of-Words (BoW) model is to describe an image with a statistical measure based on the codebook entities, e.g. a histogram describing the frequency of visual words found in the image.  The central component in this model is the codebook, which is traditionally constructed by quantizing local descriptors (e.g. SIFT) extracted from a set of training images. Consequently, to compute a representation of an image, the BoW process first determines the coding coefficients of all of the extracted local features from the image. This step involves evaluating the assignment values of the local features to the codebook entities.  Finally, the coefficients are pooled to obtain the final representation. 

%Although many such \textit{coding} and \textit{pooling} schemes have been proposed to better incorporate %the overall information content of an image, 
Before we proceed let us give a brief overview of these coding techniques, which will help us establish the mathematical framework and notation of the discussion. 

Formally, let $\mathbf{x}\in {{\mathbb{R}}^{d}}$ describe a feature vector extracted from a local patch in an image where $d$ is the dimensionality. Let $\mathbf{V}=\{{{\mathbf{c}}_{j}}\in {{\mathbb{R}}^{d}}\}_{j=1}^{K}$ be a codebook. The likelihood function $p(\mathbf{x}|\mathbf{c})$ then expresses the distribution of the local feature vector $\mathbf{x}$ given the visual word $\mathbf{c}$. Let ${{\mathbf{h}}_{i}}\in \mathcal{F}$ be the coding coefficient vector of ${{\mathbf{x}}_{i}}$ in which  a traditional BoW scheme $\mathcal{F}$ has the dimensionality equal to the cardinality of the codebook, $|\mathbf{V}|$ ($\mathcal{F} \triangleq {{\mathbb{R}}^{|\mathbf{V}|}}$). ${{h}_{ik}}$  describes the coding coefficient of ${\mathbf{x}_{i}}$ with respect to visual word ${{\mathbf{c}}_{k}}$. %After we prune a subset of visual words from an initial codebook, the resulting new coding coefficient and representation will be described by $\underline{\mathbf{h}},\underline{\mathbf{f}}\in \underline{\mathcal{F}}$, respectively. If the local features are assumed to be known, these vectors can be exactly computed thus they will represent the deterministic solution of the BoW process. Otherwise if the local features are assumed to be not known, we will regard them as random variables. This difference will be apparent or noted in the paper. $\underline{\mathcal{F}}$ now denotes the dimensionality reduced space of such coding and representation vectors.

Hard coding can then be defined as 
	\begin{equation}
{{h}_{ik}}=\left\{ \begin{matrix}
   1~~~~~~~~p\left( {{\mathbf{c}}_{k}}|\mathbf{{x}}_{i} \right)>p\left( {{\mathbf{c}}_{j}}|\mathbf{{x}}_{i} \right)~~~\forall j\ne k  \\
   0~~~~~~~~~~~~~~~~~~~~~~~otherwise~~~~~~~~~~~~~~~~  \\
\end{matrix} \right.\end{equation}
where $p\left( {{\mathbf{c}}_{k}}|\mathbf{x}_{i} \right)$ is the posterior probability of a ${{\mathbf{x}}_{i}}$ belonging to visual word ${{\mathbf{c}}_{k}}$. With $p\left( \mathbf{x}|{\mathbf{c}}_{k} \right)\sim N({\mu}_{k},{{\sigma }^{2}}I)$ and equal priors $p({\mathbf{c}}_{k})$, the condition $p\left( {{\mathbf{c}}_{k}}|{{\mathbf{x}}_{i}} \right)>p\left( {{\mathbf{c}}_{j}}|{{\mathbf{x}}_{i}} \right)~\forall j\ne k$ becomes $k=\underset{j=1,\ldots ,K}{\mathop{\arg \min }}\,\|{{\mathbf{x}}_{i}}-{{\mathbf{c}}_{j}}\|_{2}$. This assumption is considered when the codebook generation step involves $K$-means clustering. 
%When a certain visual word is omitted, say ${{\mathbf{c}}_{l}}$, this condition becomes slightly modified as $\forall j\ne k,l$. 

Similarly, soft coding can be defined as the posterior probability of assigning a local feature ${{\mathbf{x}}_{i}}$ to visual word ${{\mathbf{c}}_{k}}$,
\begin{equation}
{{h}_{ik}}=p\left( {{\mathbf{c}}_{k}}|{{\mathbf{x}}_{i}} \right)=\frac{1}{Z}\exp (-\beta \delta({{\mathbf{x}}_{i}},{{\mathbf{c}}_{k}}))
\end{equation}
where $Z=\underset{j}{\mathop \sum }\,\exp (-\beta \delta({{\mathbf{x}}_{i}},{{\mathbf{c}}_{j}}))~$is the normalization factor, $\beta$ is a parameter controlling the degree of the assignment and $\delta$ is a distance function. Finally, to obtain the BoW representation vector of the image for hard and soft coding, we consider average pooling as $\mathbf{f} =\frac{1}{N}\underset{i=1}{\overset{N}{\mathop \sum }}\,{{\mathbf{h}}_{i}}$.

In the next section we will describe our method to compute the final image representation without having to go through the usual coding and pooling steps when pruning a codeword or a subset of codewords from the codebook. 
\subsection{Methodology}
Before we proceed let us define the necessary notation. After we prune a subset of visual words from an initial codebook, the resulting new coding coefficient and representation will be described by $\underline{\mathbf{h}},\underline{\mathbf{f}}\in \underline{\mathcal{F}}$, respectively. If the local features are assumed to be known, these vectors can be exactly computed; thus, they will represent the deterministic solution of the BoW process. Otherwise we will regard them as random variables. This difference will be apparent or noted in the paper. See Table 1 for complete notation. 
\subsubsection{Hard-coding and average pooling}
After pruning a subset of visual words from the codebook, how can we infer the representation of an image with respect to the remaining codebook entities without re-coding and re-pooling? Our observation to this is that, assuming the codebook partitions the feature space with an underlying generative model, one can use this structure to infer the alteration of the representation vector without having to redo coding or pooling. To elaborate, consider the example Voronoi tesselation after quantizing a set of local features with $K$-means in Fig. 2. The Voronoi centers describe the visual words in this diagram and the Voronoi cells denote the regions consisting of local features closer (with respect to the Euclidean norm in this case) to that visual word than to any other. As a result, in hard coding, a local feature is assigned with respect to which Voronoi cell it falls into. 

Assume we prune codeword ${{\mathbf{c}}_{l}}$, the neighboring cells will then invade its region and the local features previously assigned to ${{\mathbf{c}}_{l}}$ will be distributed among them. Consider another visual word $\mathbf{c}_{k}$, the probability that a local feature previously in region of  $\mathbf{c}_{l}$ gets assigned to $\mathbf{c}_{k}$ can be approximated as a function of the location of the local feature and the proportion of volume ${{\mathbf{c}}_{k}}$ that ‘invades’ after ${{\mathbf{c}}_{l}}$ is pruned. Formally this probability can be stated as

\begin{equation}
\label{lambda}
{\Lambda}_{k,l}=\frac{1}{Z'}\underset{\underline {\mathcal{R}}({\mathbf{c}}_{k})\mathop{\cap }^{}\mathcal{R}({{\mathbf{c}}_{l}})}{\overset{{}}{\mathop \int }}\,p(\mathbf{x}|{{\mathbf{c}}_{l}})d\mathbf{x}
\end{equation}
where $Z'$ is a normalization factor described as $Z'=\underset{{{\mathbf{c}}_{k}}\in N({{\mathbf{c}}_{l}})}{\mathop \sum }\,\underset{\underline {\mathcal{R}}({\mathbf{c}}_{k})\mathop{\cap }^{}\mathcal{R}({{\mathbf{c}}_{l}})}{\overset{{}}{\mathop \int }}\,p(\mathbf{x}|{{\mathbf{c}}_{l}})d\mathbf{x}$ in which $\mathcal{R}({\mathbf{c}}_{k})$ and $\underline {\mathcal{R}}({\mathbf{c}}_{k})$ denote the regions in the descriptor space where $p\left( \mathbf{x}|{{\mathbf{c}}_{k}} \right)>p\left( \mathbf{x}|{{\mathbf{c}}_{j}} \right)~\forall j\ne k$ before and after removing visual word ${\mathbf{c}}_{l}$, $\mathop{\cap }^{}$ describes the intersection of the regions and $N({\mathbf{c}}_{l})$ is the set of visual words for which their regions are incident to $\mathcal{R}({\mathbf{c}}_{l})$.

After omitting ${\mathbf{c}}_{l}$, let the new image representation obtained through the usual BoW process be $\underline{\mathbf{f}}=\frac{1}{N}\underset{i=1}{\overset{N}{\mathop \sum }}\,{\underline{{\mathbf{h}}_{i}}}$ where $\underline{{\mathbf{h}}_{i}}, \underline{{\mathbf{f}}}\in \underline{\mathcal{F}}$ and let $\mathcal{I}=\{1,..,K\}$ denote the visual word indices in the initial codebook. We then specify a mapping ${\Psi}:\mathcal{F}\times I\to \underline{\mathcal{F}}$ to approximate $\underline {\mathbf{f}}$. This mapping function is described as 
\begin{equation}
\label{hardcoding}{{\psi }_{k}}\left( \mathbf{f},l \right)=\left\{ \begin{matrix}
   {{f}_{k}}+{\Lambda}_{k,l}\times {{f}_{l}} ~~~~~~~~~{{\mathbf{c}}_{k}}\in N\left( {{\mathbf{c}}_{l}} \right),~k<l  \\
   {{f}_{k}}~~~~~~~~~~~~~~~~~~~~~~~~~~~{{\mathbf{c}}_{k}}\notin N\left( {{\mathbf{c}}_{l}} \right),~k<l  \\
\end{matrix} \right.
\end{equation}
where $l$ denotes the index of the omitted visual word in the codebook\footnote{Without loss of generality we assumed $l = max(I)$.}. If we consider the conditional densities to be Gaussian distributions, $p\left( \mathbf{x}|{\mathbf{c}}_{k} \right)\sim{\ }N({{\mu}_{k}},{{\Sigma}_{k}})$, then the regions correspond to the Voronoi cells under the decomposition of a Mahalanobis metric space. 
 \begin{figure}[t]
  \centering
    \includegraphics[width=0.30\textwidth]{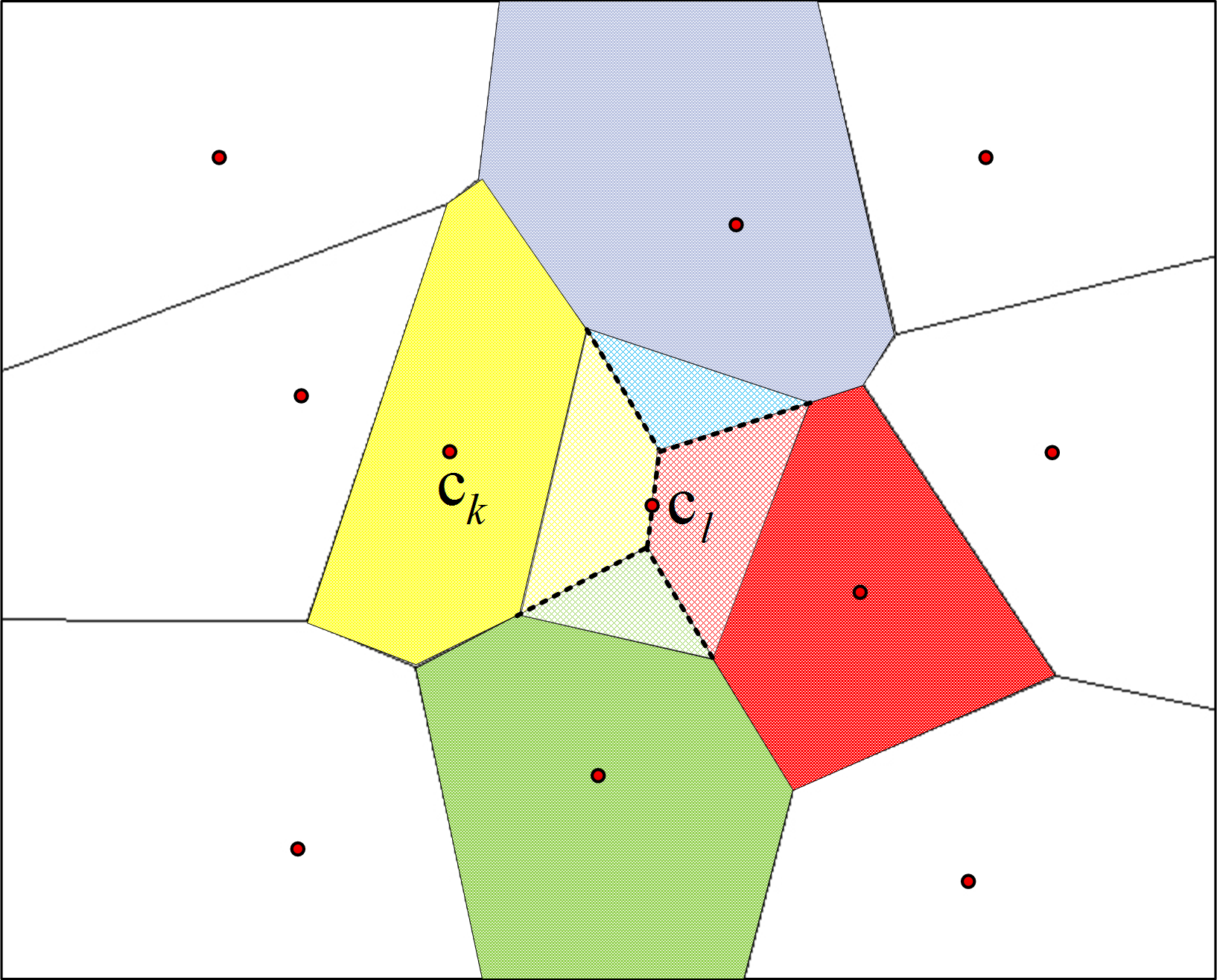}
    \caption{Voronoi tesselation of the feature space describing the quantization of a set of local descriptors. The Voronoi centers denote the visual words and Voronoi cells denote the regions consisting of local descriptors closer to that visual word than to any other. }
\end{figure}

\noindent \textbf{Proposition 1 (Hard-coding and average pooling)} Under hard coding and average pooling, when local features are i.i.d sampled from $p(\mathbf{x}|{{\mathbf{c}}})$, then $\mathbb{E}[\underline{\mathbf{f}}]=\Psi \left( \mathbf{f}, l \right)$. 

\noindent \textbf{Proof:} Let $\underline{{f}_{k}}=\frac{1}{N}\underset{i=1}{\overset{N}{\mathop \sum }}\,\underline{{h}_{ik}}=\frac{1}{N}[\underset{i\in {{S}_{1}}}{\mathop \sum }\,\underline{{h}_{ik}}+\underset{i\in {{S}_{2}}}{\mathop \sum }\,\underline{{h}_{ik}}]$ where we partition the local features into two disjoint sets ${{S}_{1}}$ and ${{S}_{2}}$, where ${{S}_{1}}$ contains the indices of the local features that are $p\left( {{\mathbf{c}}_{l}}\text{ }\!\!|\!\!\text{ }{{\mathbf{x}}_{i}} \right)>p\left( {{\mathbf{c}}_{k}}\text{ }\!\!|\!\!\text{ }{{\mathbf{x}}_{i}} \right)~\forall k\ne l$ and ${{S}_{2}}$ contains rest of the indices. Thus the coding coefficients $\underline{{h}_{ik}}~(i\in {{S}_{1}})$ are Bernoulli random variables while $\underline{{h}_{ik}}~(i\in {{S}_{2}})$ are deterministic. We see that $\frac{1}{N}\underset{i\in {{S}_{2}}}{\mathop \sum }\,\underline{{h}_{ik}}=\frac{1}{N}\underset{i=1}{\overset{N}{\mathop \sum }}\,{{h}_{ik}}$ and thus ${\mathbb{E}}[ \frac{1}{N}\underset{i\in {{S}_{2}}}{\mathop \sum }\,\underline{{h}_{ik}}]={{f}_{k}}$. 

If  ${{\mathbf{c}}_{k}}\in N({{\mathbf{c}}_{l}})$ then $\Pr [\underline{{h}_{ik}}=1]={\Lambda}_{k,l}~(i\in {{S}_{1}})$ where ${\Lambda}_{k,l}$ is defined as above. %\footnote{Intuitively this is because, after pruning the visual word ${{\mathbf{c}}_{l}}$ its region will be ‘invaded’ by its neighbour regions, $N({{\mathbf{c}}_{l}})$. The probability of a local feature being in $\mathcal{R}({{\mathbf{c}}_{k}})$ is a function of the location of the local feature and the proportion of volume ${{\mathbf{c}}_{k}}$ ‘invades’ after ${{\mathbf{c}}_{l}}$ is pruned.},
Consequently, 
\begin{equation}
\label{proposition1proof}
\begin{split}
\mathbb{E}[ \underline{{f}_{k}} ] &=\mathbb E\left[ \frac{1}{N}[ \underset{i\in {{S}_{1}}}{\mathop \sum }\,\underline{{h}_{ik}}+\underset{i\in {{S}_{2}}}{\mathop \sum }\,\underline{{h}_{ik}} ]\right] \\
&=\frac{1}{N} \mathbb E[ \underset{i\in {{S}_{1}}}{\mathop \sum }\,\underline{{h}_{ik}} ]+{{f}_{k}} \\
&=\frac{{\Lambda}_{k,l}\times |{{S}_{1}}|}{N}+{{f}_{k}}
\end{split}
\end{equation} where $\frac{|{{S}_{1}}|}{N}={{f}_{l}}$ and $ \mathbb E[\underset{i\in {{S}_{1}}}{\mathop \sum }\,\underline{{h}_{ik}}]$ is the expected value of sum of Bernoullis, a Binomial distribution. Note also that $\text{var}\left( \underline{{f}_{k}} \right)=\frac{\left| {{S}_{1}} \right|{{\Lambda}_{k,l}}\left( 1-{{\Lambda}_{k,l}} \right)}{{{N}^{2}}}$.

If ${{\mathbf{c}}_{k}}\notin N({{\mathbf{c}}_{l}})$ then $\Pr \left[ \underline{{h}_{ik}}=1 \right]=0~(i\in {{S}_{1}})$ since $\mathcal{R}\left( {{\mathbf{c}}_{k}} \right)\mathop{\cap }^{}\mathcal{R}\left( {{\mathbf{c}}_{l}} \right)=\emptyset$. This is true because if the pruned visual words ${{\mathbf{c}}_{l}}$ and ${{\mathbf{c}}_{k}}$ are not neighbours in the Voronoi tessellated feature space then after pruning 
${{\mathbf{c}}_{l}}$, $\exists j$ such that $p\left( {{\mathbf{c}}_{j}}\text{ }\!\!|\!\!\text{ }{{\mathbf{x}}_{i}} \right)>p({{\mathbf{c}}_{k}}|{{\mathbf{x}}_{i}})$. Then $\mathbb E[\underset{i\in {{S}_{1}}}{\mathop \sum }\,\underline{{h}_{ik}}] = 0$, and thus $\mathbb{E}\left[ \underline{{f}_{k}} \right] = {f}_{k}$. Finally, we see that $\mathbb{E}\left[ \underline{{f}_{k}} \right]={\psi}_{k} \left( \mathbf{f},l \right)$.
%, hence by Monte Carlo principle $\underset{N\to \infty }{\mathop{\lim }}\,\underline{\mathbf{f}}=\Psi \left( \mathbf{f}, l \right)$. 
\begin{flushright} 
$\square$ 
\end{flushright}

The above proposition is important, as it states that given an image, with hard coding and average pooling, if the local features are sampled densely enough (or the number of local features is high) Eq. \ref{hardcoding} allows one to compute the representation of an image without having to do re-coding and pooling, which is a significant advantage since it bypasses the computationally expensive nearest neighbour searches. 

%Two difficulties arise in this hard coding approximation case. The first is, we assumed $\mathcal{F} \triangleq{{\mathbb{R}}^{k}}$ and ${\underline{\mathcal{F}}\triangleq{\mathbb{R}}^{k-1}}$, i.e., our analysis only considers the case when a single visual word is omitted. What if we would like to infer the altered representation when multiple visual words are omitted at once? We leave this problem as future work but our solution to this problem is to apply the proposed approximation scheme in a certain order.
%\footnote{In the appendix, we will give the more general solution when $\underline{\mathcal{F}}\triangleq{{\mathbb{R}}^{m}}$ where $m < K$, but I believe even if apply the proposed approximation scheme of Eq. 2 in any arbitrary order we will get the same solution.}. 
%The required calculations of the Voronoi regions in a %n usually
 %high-dimensional descriptor space are computationally expensive. 
If we consider $p\left( \mathbf{x}|{\mathbf{c}}_{k} \right)\sim{\ }N({{\mu }_{k}},{{\sigma }^{2}}I)$ in which the codebook corresponds to the Voronoi tesselation under the Euclidean metric, the complexity of computing the Voronoi diagram in $d$-dimensions is computationally prohibitive \cite{Komei2012}.
%$O({{n}^{\frac{d+1}{2}+1}}d)$\footnote{I am not sure about this. From http://www.ifor.math.ethz.ch/~fukuda/polyfaq/node31.html.}. 
In the general case, we propose to use a heuristic, as explained next, which uses nearest neighbor information of a visual word to approximate $\underline{\mathcal{R}}({{\mathbf{c}}_{k}})\mathop{\cap }^{}\mathcal{R}({{\mathbf{c}}_{l}})$ in Eq.~\ref{lambda}. 
\\
\\
\noindent \textbf{Heuristic for hard-coding}\label{sec:heuristics}
%As stated, one has to construct the Voronoi tessellation of the codebook before and after the visual word ${{\mathbf{c}}_{l}}$ is pruned to evaluate Eq. \ref{lambda}. However, 
If the dimensionality of the feature space is high, constructing the Voronoi diagram to evaluate Eq. \ref{lambda} is generally intractable .
%Although the resulting likelihood functions are known for popular clustering algorithms such as $K$-means and Expectation-Maximization for GMMs, 
In order to also avoid the integration in high dimensions of Eq. \ref{lambda} we only use the neighborhood information of the pruned visual word to approximate ${\Lambda}_{k,l}$. This neighborhood information is with respect to a metric, not the neighbors in the Voronoi diagram of the pruned visual word ${{\mathbf{c}}_{l}}$ $(N({{\mathbf{c}}_{l}}))$. We show that this simplification shows good performance results. 

Formally, we revise Eq. \ref{hardcoding} to incorporate this heuristic. Assume the set ${{N}_{\delta}}({{\mathbf{c}}_{l}})$ contains the nearest neighbors of ${{\mathbf{c}}_{l}}$ in the codebook space based on some distance function $\delta:\mathcal{X}\times \mathcal{X}\to \mathbb{R}$. For example, if the initial codebook has been constructed by $K$-means this distance function would be the ${{l}_{2}}-$norm. The cardinality of this set is a user specified parameter. Consequently, Eq. \ref{hardcoding} is revised as 
\begin{equation}
\label{hardcodingheuristic}
	{\psi}_{k} \left( \mathbf{f},l \right)=\left\{ \begin{matrix}
   {{f}_{k}}+\frac{{{f}_{l}}}{|{{N}_{\delta}}({{\mathbf{c}}_{l}})|}~~~~~~~~{{\mathbf{c}}_{k}}\in {{N}_{\delta}}\left( {{\mathbf{c}}_{l}} \right),~k<l  \\
   {{f}_{k}}~~~~~~~~~~~~~~~~~~~~~~~~~~{{\mathbf{c}}_{k}}\notin {{N}_{\delta}}\left( {{\mathbf{c}}_{l}} \right),~k<l  \\
\end{matrix} \right.
\end{equation}
\subsubsection{Soft-coding and average pooling}
In this section we will examine another popular coding scheme used in the literature. Compared to hard coding, soft coding assigns a local feature to each codebook entity and computes its degree of membership to them. When a visual word is pruned, the new representation vector can be computed exactly if we maintain the initial coding information of the local features. Formally, let $\mathcal{I}=\{1,..,K\}$ again denote the visual word indices in the initial codebook.   With soft-coding and average pooling schemes, assume we maintain the initial coding vectors $\mathbf{H}=[{{h}_{ik}}]$ where ${{h}_{ik}}=\exp (-\beta \delta \left( {{\mathbf{x}}_{i}},{{\mathbf{c}}_{k}} \right))/\underset{j}{\mathop \sum }\,\exp (-\beta \delta \left( {{\mathbf{x}}_{i}},{{\mathbf{c}}_{j}} \right))$. Then we describe the new feature representation as
	\begin{equation}
\label{claim2}{{\upsilon }_{k}}\left( \mathbf{H},l \right)=
  \frac{1}{N} \underset{i=1}{\overset{N}{\mathop \sum }}\,[ \frac{{{h}_{ik}}}{\mathop{\sum }_{k=1,k\notin S}^{K}{{h}_{ik}}} ]
\end{equation}
where $\mathcal{S} \subset \mathcal{I}$ contains the indices of the visual words to be pruned. Note that $\underline{{f}_{k}}=\frac{1}{N}\underset{i=1}{\overset{N}{\mathop \sum }}\,\underline{{h}_{ik}}$ where $\underline{{h}_{ik}}=\exp \left( -\beta \delta \left( {{\mathbf{x}}_{i}},{{\mathbf{c}}_{k}} \right) \right)/\underset{k=1,k\notin S}{\overset{K}{\mathop \sum }}\,\exp (-\beta \delta ({{\mathbf{x}}_{i}},{{\mathbf{c}}_{k}}))$ represents the deterministic (exact) solution of the BoW model when the subset of visual words are pruned. 
\\
\\
\noindent \textbf{Claim 2 (Soft-coding and average pooling)} Under such schemes, $\boldsymbol{\upsilon}=\underline {\mathbf{f}} .$

\noindent \textbf{Proof: } 
It is easy to verify that
	\begin{equation}
\label{claim2proof}
\begin{split}
{{v}_{k}}=\frac{1}{N}\underset{i=1}{\overset{N}{\mathop \sum }}\,\frac{{{h}_{ik}}}{\left[ \mathop{\sum }_{k=1,k\notin S}^{K}{{h}_{ik}} \right]}&=\frac{\frac{\exp \left( -\beta \delta \left( {{\mathbf{x}}_{i}},{{\mathbf{c}}_{k}} \right) \right)}{\mathop{\sum }_{k=1}^{K}\exp \left( -\beta \delta \left( {{\mathbf{x}}_{i}},{{\mathbf{c}}_{k}} \right) \right)}}{\frac{\mathop{\sum }_{k=1,k\notin S}^{K}\exp \left( -\beta \delta \left( {{\mathbf{x}}_{i}},{{\mathbf{c}}_{k}} \right) \right)}{\mathop{\sum }_{k=1}^{K}\exp \left( -\beta \delta \left( {{\mathbf{x}}_{i}},{{\mathbf{c}}_{k}} \right) \right)}}\\
&=\underline{{f}_{k}}
\end{split}
\end{equation}
\begin{flushright} $\square$ \end{flushright}
Thus, maintaining the initial coding vectors allows us to exactly compute the new representation vector.  %Note that $\mathbf{H}$ represents the matrix of coding coefficients of a single image, not the entire dataset; thus, it does not require heavy memory usage. The coding matrices correspoding to all images can be computed offline and stored. 
%\subsubsection{Fisher Vector Coding}
%In this section we will examine a recent enconding scheme that produces the state-of-the-art results among BoW models \cite{devil2011}.
\subsection{Example Codeword Selection Application}
%In the BoW framework, the cardinality of the codebook is important since the dimensionality of the feature representations depends on it. 
%As stated earlier, many codeword selection methods have been introduced to reduce the dimensionality by selecting visual words most informative to the task at hand. However this reduction is performed after computing the representation vectors with respect to the initial codebook due to the fact that the analysis for which codeword to prune is generally made with respect to the initial codebook and representations.
We now describe an example codeword selection scheme, where a subset of codewords is selected to maximize a given objective function via a simulated annealing process. The techniques developed in the previous section allow us to bypass the computationally expensive re-coding step (and also re-pooling for hard coding), which allows us to compute the altered representations under various subsets of visual words very efficiently. %One alternative way to reduce the dimensionality \textit{and} to eliminate the requirement of computing the coding coefficient with respect to the initial vocabulary is to construct one with a reduced size, but this gives us a new basis that generally is found by minimizing a certain reconstruction error. Even if the clustering mechanisms involve a more complex criterion the clustering process still may take a considerable amount of time. Thus, in this section a codeword selection method is proposed based on simulated annealing process.
 \begin{algorithm}[h!]
 \hrulefill
 \SetAlgoLined
 
 \KwData{Initial codebook $V=\{{{\mathbf{c}}_{1}},{{\mathbf{c}}_{2}},\ldots ,{{\mathbf{c}}_{k}}\}$, initial representation matrix of the data ${\Pi}^{0}$, $tmax, \lambda$ } 
 \Begin{
 ${\mathcal{T}}^{0} \leftarrow$ initial subset of visual words (initial state)\;
 ${e}^{0} \leftarrow$ initial energy\;
 $t\leftarrow0$ \;
 \While{t $<$ tmax}{
  compute altered feature representation according to Eq. \ref{hardcodingheuristic} or Eq. \ref{claim2}\;
  $\underline{\Pi^{t}}\leftarrow get\_feature\_representations(\Pi^{t-1},{\mathcal{T}}^{t})$\;
  assuming a distribution for each bin (visual word ${v}_{k}$), estimate its parameters\;
  $\{\hat \theta \}_{i=1}^{|{\mathcal{T}}^{t}|} \leftarrow maximum\_likelihood(\underline{\Pi^{t}})$\;
  calculate score according to Eq. \ref{score}\;
  ${e}^{t}\leftarrow maximum\_relevance(\{\hat \theta \}_{i=1}^{|{\mathcal{T}}^{t}|},\underline{\Pi^{t}})$\;
  $\Delta e = {e}^{t} - {e}^{t-1}$\;
  $\Pr \left[ Accept \right]\leftarrow{{e}^{\frac{\Delta e}{{{\lambda}^{t}}}}}$\;
   \If(\tcc*[f]{Reject, ${\mathcal{T}}^{0}$ always accepted}){$\Pr \left[ Accept \right] < random(0,1)$}{
   ${e}^{t} \leftarrow {e}^{t-1}$\;
   ${\mathcal{T}}^{t} \leftarrow {\mathcal{T}}^{t-1}$\;
  }
  get neighbour state\;
  ${\mathcal{T}}^{t+1}\leftarrow neighbour({\mathcal{T}}^{t})$\;
  $t\leftarrow t+1$\;
 }
 }
 \KwResult{${\mathcal{T}}^{t}$}
 \hrulefill
 \caption{Simulated annealing based codeword selection.}
\end{algorithm}

There are many scoring functions, such as error probability, inter-class distance, etc., to employ in a standard feature selection technique. After obtaining the final representation vectors (i.e. histogram) of images in the dataset, $\underline{\mathbf{{f}}}$, we consider the maximum relevancy score \cite{max2005} defined as 
\begin{equation}
\label{score}
D\left( \mathcal{T},y\right)=\frac{1}{|\mathcal{T}|}\underset{\underline{{{f}_{k}}}\in \mathcal{T}}{\mathop \sum }\,I(\underline{{{f}_{k}}};y)
\end{equation}
where $\mathcal{T}$ is the set of indices of the visual words used to infer the representation vector (i.e., the set $\mathcal{I}\backslash \mathcal{S}$)\footnote{In which $\mathcal{S}$ contains the indices of the visual words that are pruned.} and $I\left( \underline{{{f}_{k}}};{y} \right)$ is the mutual information defined as a function of the (differential) entropy with respect to the ${k}^{th}$ bin of $\underline{{f}}$ and class $y$:
\begin{equation}
I\left( \underline{{{f}_{k}}};y \right)=h\left( \underline{{{f}_{k}}} \right)-\mathop{\int }^{}p\left( y \right)h(\underline{{{f}_{k}}}|y)dy.
\end{equation}

We summarize the feature selection procedure in Algorithm 1 in which we assume $\underline{{{f}_{k}}}\sim\text{Beta}(\alpha ,\beta )$ where the parameters are estimated via maximum likelihood given the feature representation vectors of the data. 
\begin{figure*}[t]
  \centering
  \label{figur}
    \subfloat[Hard coding - 15 Scenes - 1000]{\label{figur:2}\includegraphics[width=42mm]{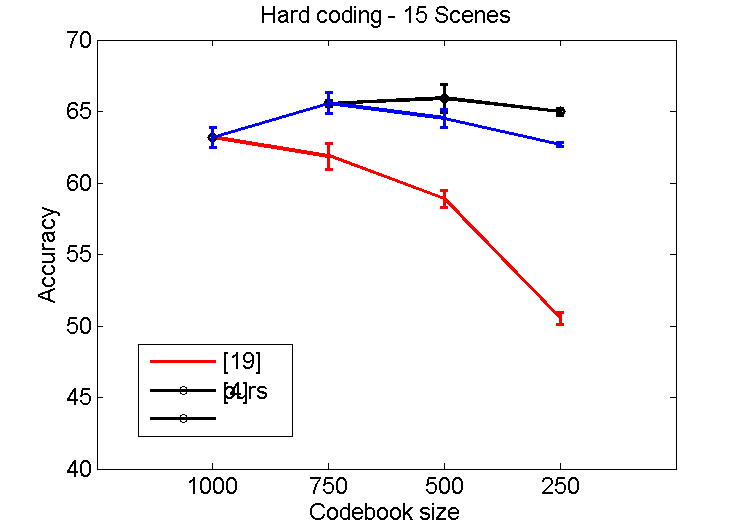}}
  \subfloat[Hard coding - Caltech 10 - 1000]{\label{figur:6}\includegraphics[width=42mm]{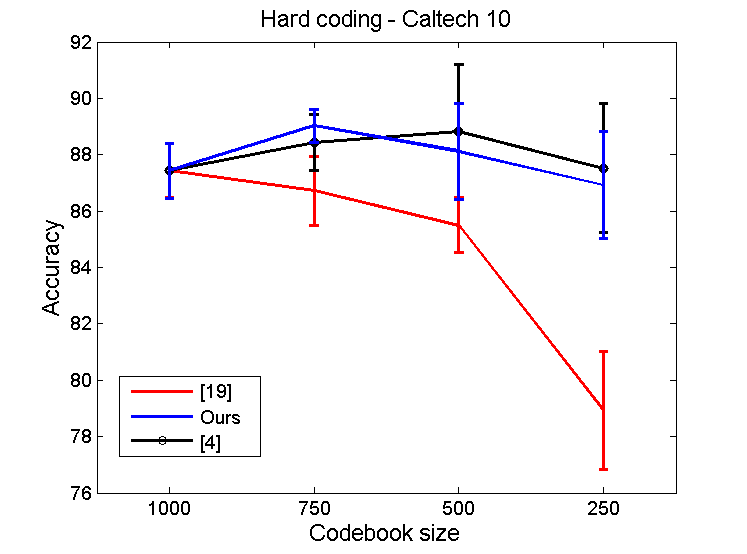}} 
   \subfloat[Hard coding - 15 Scenes - 5000]{\label{figur:5}\includegraphics[width=42mm]{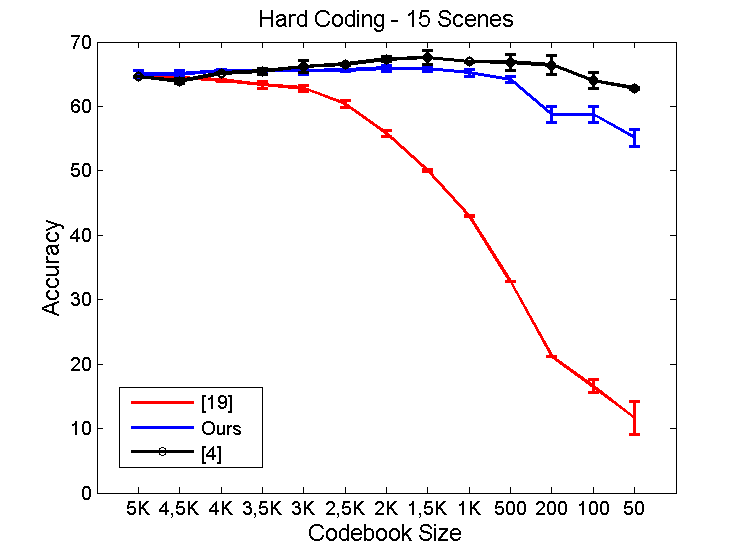}}
  \subfloat[Hard coding - Caltech 10 - 5000]{\label{figur:7}\includegraphics[width=42mm]{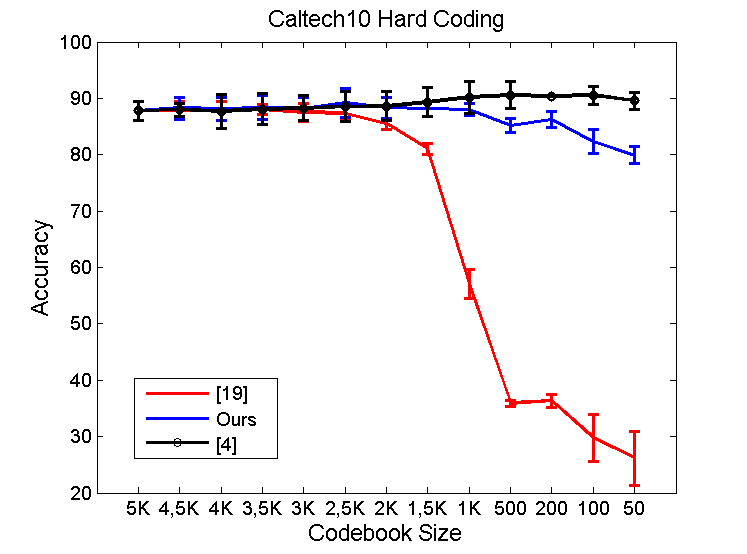}}\\
  \subfloat[Soft coding - 15 Scenes - 1000]{\label{figur:4}\includegraphics[width=42mm]{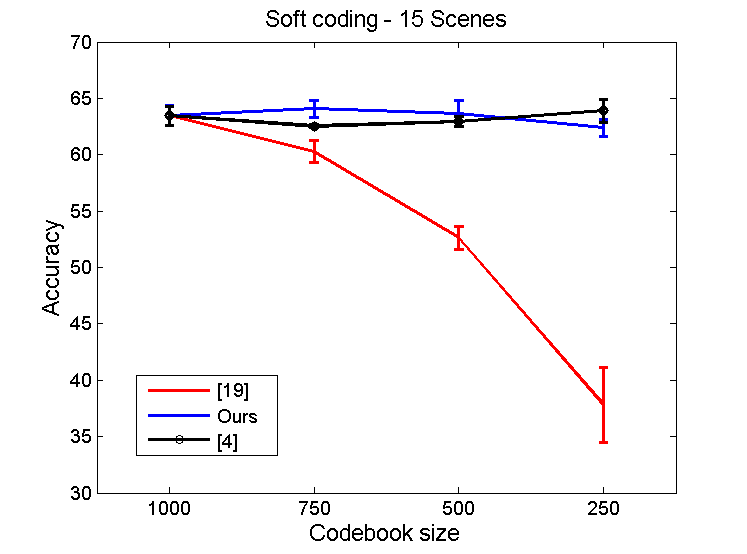}}
  \subfloat[Soft coding - Caltech 10 - 1000]{\label{figur:8}\includegraphics[width=42mm]{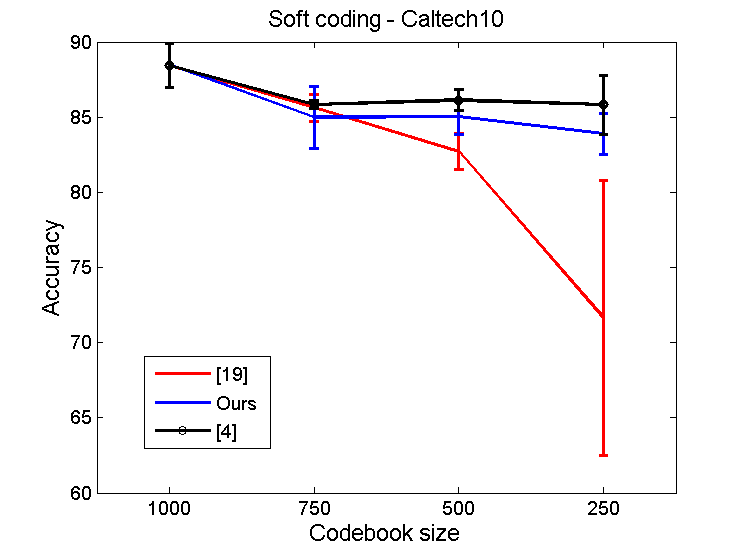}}
  \subfloat[Soft coding - 15 Scenes - 5000]{\label{figur:6}\includegraphics[width=42mm]{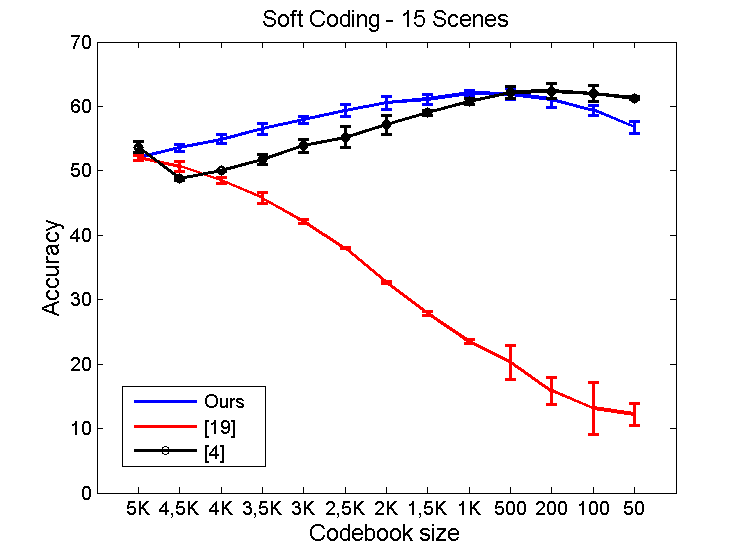}} 
  \subfloat[Soft coding - Caltech 10 - 5000]{\label{figur:8}\includegraphics[width=42mm]{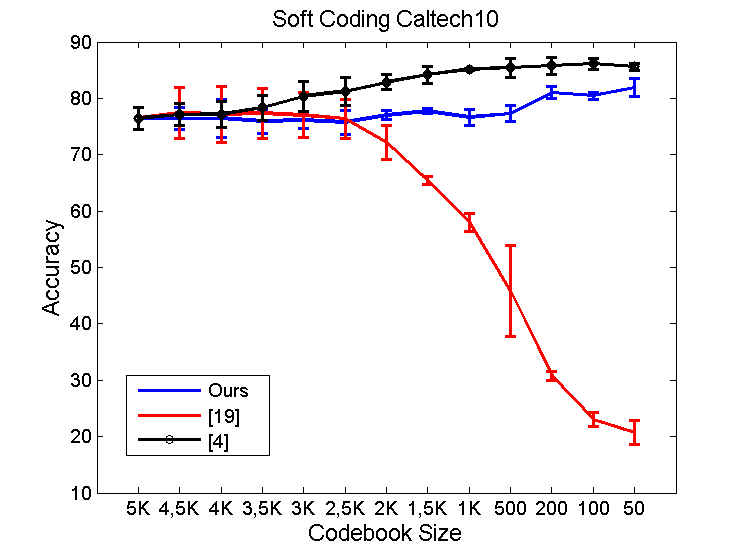}}
\caption{\label{fig:graphs}Comparison of our method with respect to a baseline. The vertical axes in the sub-figures denote the average classification accuracy of the categories while the horizontal axes denote the codebook sizes. The caption 'Soft coding - Caltech 10 - 1000' denotes the soft coding case performed on the Caltech 10 dataset with an initial codebook size of 1000.}
\end{figure*}
\section{Experiments}
In our experiments, our focus is on not to achieve state-of-the-art classification results but to demonstrate that the techniques we developed can be used in a codeword selection problem where the task is to obtain a compact codebook. %via pruning visual words while eliminating the expense of re-coding vectors with respect to the initial codebook. %We conduct experiments to evaluate the performance of Alg.~1. 
%In this case we demonstrate that given a codebook generated via $K$-means, our pruning strategy yields a compact codebook that performs at least competitively compared to the traditional codeword selection schemes with the difference that our method eliminates the need to compute coding vectors with respect to the initial codebook.
We compare our method against \cite{yang2007,fulkerson}, where similar entropy-based measures are used for characterizing the discriminative power of the visual words.  
We hope to observe little sacrifice in classification performance vs.\ these techniques, while avoiding the expense of re-coding representation vectors with respect to an initial codebook.

%; ii) for hard coding, the approximated representations when a subset of visual words is pruned demonstrates similar performances in a classification setting compared to the case in which we use the complete BoW process to compute such representations.

Two image classification benchmark datasets are used in our evaluation: 15-Scenes and Caltech-10. For both datasets, each image is first resized such that neither its height nor width exceeds 300 pixels. Densely sampled SIFT \cite{sift1999} descriptors on a single scale of $16\times 16$ patches with step size equal to 8 pixels are extracted. We use $K$-means clustering to create two visual vocabularies for each dataset, where the number of clusters is set to 1000 and 5000. The number of neighbors that determines the cardinality of $|{{N}_{\delta}}\left( {{\mathbf{c}}_{l}} \right)|$ is set to be 5. $\lambda $ is set to be $0.9$ and $tmax$ is set to $100$ and $500$ for the hard coding and soft coding cases, respectively. 
%The value of is $tmax$ chosen to be smaller in the hard coding case since Eq.~\ref{hardcodingheuristic} is an approximation and with larger step sizes the representation vector becomes too randomized. 
The neighbor state in Alg.~1 alters set $T$ by replacing a subset with their corresponding nearest neighbors in codebook space. The size of this subset is chosen to be 10. Finally all experiments are conducted 5 times over random subsets of images. 

For 15-Scenes, 50 images are used for generating a codebook and an additional 50 images are used for determining which visual word to prune based on Alg.~1. Once the visual words are pruned, the same 50 images are used for training a linear SVM. Hence, 100 images are used in total for training, and the remaining images are used for testing. 

Our second benchmark contains the largest 10 categories (except for BACKGROUND\_Google category and Faces\_easy) from the Caltech-101 dataset. 25 images per category are used to train a codebook and 25 are used to select visual words to prune based on Alg.~1. Once the visual words are pruned, the same 25 images are used for training a linear SVM. The  remaining images are used for testing.

\subsection{Discussion}
Fig.~\ref{fig:graphs} shows our results. The vertical axes in the sub-figures denote the average classification accuracy of the categories while the horizontal axes denote the codebook sizes. %First of all, the variances of the performances, especially for the hard coding case are high in certain cases. This is not surprising since compared to the experiments based on evaluating Soft coding, our technique to infer the final representation for the Hard coding case is based on a heuristic. 
First of all we see that the approach of \cite{yang2007} shows inferior performance compared to \cite{fulkerson} and our technique. This is especially apparent when the initial codebook size is large.  After a certain number of visual word pruning steps, the classification performance degrades severely. This suggests that removing the bins of the image representation corresponding to the pruned set of visual words also reduces the discriminative power of the representation. On the other hand, the performance of \cite{fulkerson} shows that merging the bins instead of completely discarding them can improve the classification performance. This may be due to the fact that certain visual words correspond to the same texture but due to different lighting conditions or other effects it may have been represented by multiple words. Hence, merging the bins will not reduce the discriminative ability but enhance it while also eliminating the redundancy in the codebook.
 
Our technique shows competitive performance with \cite{fulkerson}. Notice that for both techniques the curve of the accuracy values across different sizes of the codebook demonstrate that under certain tasks the codebook is overcomplete with noisy visual words. In fact in certain cases we decrease the size of the codebook by two orders of magnitude while increasing the overall accuracy. Also we see that both \cite{fulkerson} and our technique have more impact when started with an overcomplete codebook. One other thing to notice that both hard and soft coding outperformed each other in certain cases. %but it not only reduces the dimensionality of the representation, it also \textit{truly} eliminates requirement of doing coding and pooling . 

However, the crucial advantage of our method is the fact that we truly reduce the size of the codebook and thus need not store it. We explore the space of subsets of visual words via the Simulated Annealing method. A disadvantage of this method is that it may require lengthy simulations to find a ``good'' state, i.e. the subset of visual words that is satisfactory, since the number of states to consider is very large and also the energy function is multimodal. Despite these issues, our method performs well in relatively few iterations as noted in the preceding section. Moreover, this exploration would have normally required the instantiation of the BoW process for each subset of the codebook, but our analysis alleviates this burden by approximating the image representations without doing coding and pooling.

Overall, compared to a brute-force solution of our problem, we explored the solution space (subset of visual words) without having to do re-coding and re-pooling. Compared to previous codeword reduction techniques this analysis enables us to truly reduce the codebook. %size by eliminating the need for coding with respect to the initial words. 

\subsubsection{Complexity Analysis}
%We now analyze the reductions in complexity afforded by our formulation.% by not doing re-coding and/or re-pooling.
Compared to previous codeword selection schemes, we reduce the time complexity by eliminating the need of re-coding with respect to the initial codebook. For example, for the Caltech 10 dataset, \cite{fulkerson} reduces the size of the codebook by two orders of magnitude, from 5000 to 50. However, they still require doing coding with respect to the 5000 visual words to compute image representations, which is computationally inefficient if one considers thousands of images. Formally, after the codeword selection process. the complexity of coding new image representations becomes $\mathcal{O}(d|\mathcal{T}|N)$ compared to $\mathcal{O}(d|V|N)$ where $V$ denotes the initital codebook and $|\mathcal{T}| << |V|$. Likewise we also reduce the space complexity since we do not need to store the 5000-word codebook. 

As stated, for each state during the simulation, the BoW must be instantiated for an entire corpus. However, we explore the state space without having to do re-coding and re-pooling. Compared to the brute-force approach where the BoW is instantiated at every step, we analyze the reductions in complexity afforded by our formulations.
\begin{figure}[t!]
  \centering
    \includegraphics[width=0.4\textwidth]{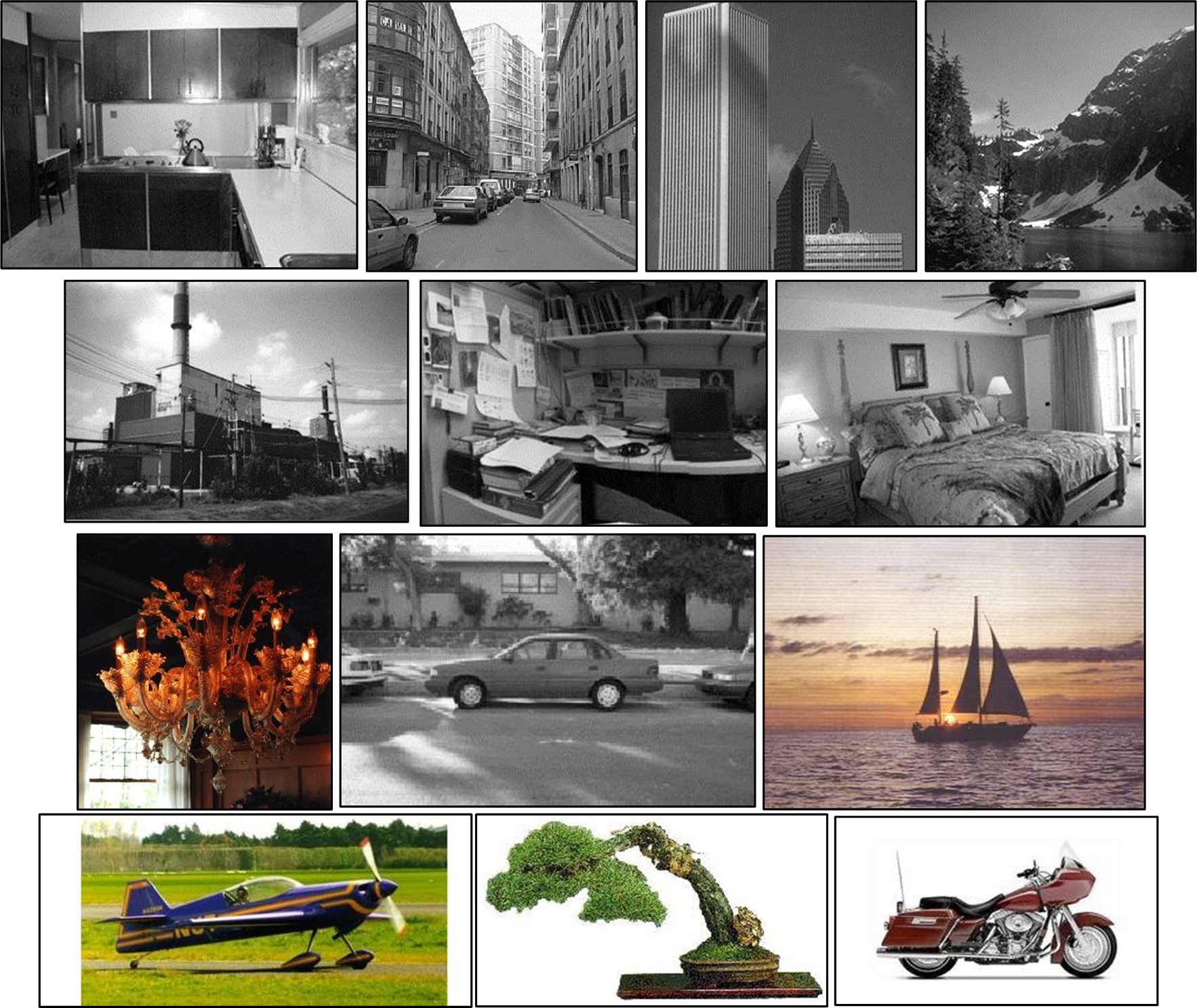}
    \caption{Example images from the Caltech 10 (top two rows) and 15 Scenes datasets (bottom two rows).}
\end{figure}

\noindent \textbf{Hard-coding:}
Suppose we remove the set of visual words $\mathcal{S}$, instantiating the BoW pipeline to compute the new representation under the set of the remaining visual words $\mathcal{T}$ would require a nearest-neighbor search for all extracted local features. Given the fact that a na\"{i}ve NN search outperforms techniques based on partitioning the feature space in high dimensions, we assumed an $\mathcal{O}(d~|\mathcal{T}|)$ complexity \cite{nnsearch}. Thus, the complexity is $\mathcal{O}(d~|\mathcal{T}|~N)$ where $N$ is the number of local features with $d$ dimensionality.  This coding step dominates pooling, which has complexity $\mathcal{O}(|\mathcal{T}|~N)$. 

In contrast, our technique has $\mathcal{O}(|\mathcal{S}||{N}_{\delta}(\mathbf{c}_{l})|)$ complexity where $|{N}_{\delta}(\mathbf{c}_{l})| << |V|$. For each deleted visual word we distribute its bin value among its $|{N}_{\delta}(\mathbf{c}_{l})|$ neighbors. This neighbor information can be computed offline. 

\noindent \textbf{Soft-coding:} Instantiating the BoW pipeline for computing the image representation under the set of remaining visual words again requires $\mathcal{O}(d~|\mathcal{T}|~N)$. However, maintaining the initial coding matrix $\mathbf{H}$ allows us to reduce this complexity to $\mathcal{O}(N|\mathcal{T}|)$.  

\section{Conclusion}
We formulated and analyzed a method for inferring the representation vector in an assignment based BoW model, without the need to re-code or re-pool as visual words are pruned from a vocabulary. The formulation is demonstrated in an  efficient and effective simulated annealing scheme that prunes words from a codebook. Compared to similar entropy-based solutions, our algorithm demonstrates superior results to \cite{yang2007}, and roughly comparable results to \cite{fulkerson} but enables reduced time and space complexity for subsequent BoW representations.  We expect that our basic strategy should be applicable to other assignment based BoW models, such as the super vector \cite{sv2010}, Fisher \cite{fish2010} and VLAD \cite{vlad} encoding methods. This will be explored in future work.
%Since these are all based on hard and soft coding schemes, the extensions will also be primarily based on techniques studied in this paper and thus not be very difficult. %Another possible application would be of finding the optimal codebook size which is also important in recognition problems. For this task, normally one would have to construct vocabularies with varying sizes and perform experiments on them, however our method opens a way to eliminate words by enabling the analysis of the representation vectors to be done very efficiently.

{\small
\bibliographystyle{ieee}
\bibliography{egbib}
}
\end{document}